\documentclass[conference]{IEEEtran}
\IEEEoverridecommandlockouts
\usepackage{amsmath,amssymb,amsfonts}
\usepackage{algorithmic}
\usepackage{graphicx}
\usepackage{textcomp}
\usepackage{cleveref}
\usepackage{xcolor}
\usepackage{biblatex}
\usepackage{array}
\usepackage{subcaption}
\bibliography{refs}
\newcolumntype{P}[1]{>{\centering\arraybackslash}p{#1}}
\def\BibTeX{{\rm B\kern-.05em{\sc i\kern-.025em b}\kern-.08em
    T\kern-.1667em\lower.7ex\hbox{E}\kern-.125emX}}
\DeclareUnicodeCharacter{0301}{é}
\begin{document}

\title{Robotic Arm Manipulation to Perform Rock Skipping in Simulation
}

\author{\IEEEauthorblockN{Nicholas Ramirez \& Michael Burgess}
\IEEEauthorblockA{\textit{Department of Electrical Engineering and Computer Science (EECS)} \\
\textit{Massachusetts Institute of Technology (MIT)}\\
Cambridge, MA, USA}
}

\maketitle

\begin{abstract}
Rock skipping is a highly dynamic and relatively complex task that can easily be performed by humans. This project aims to bring rock skipping into a robotic setting, utilizing the lessons we learned in Robotic Manipulation. Specifically, this project implements a system consisting of a robotic arm and dynamic environment to perform rock skipping in simulation. By varying important parameters such as release velocity, we hope to use our system to gain insight into the most important factors for maximizing the total number of skips. In addition, by implementing the system in simulation, we have a more rigorous and precise testing approach over these varied test parameters. However, this project experienced some limitations due to gripping inefficiencies and problems with release height trajectories which is further discussed in our report. 
\end{abstract}

\begin{IEEEkeywords}
rock, skipping, manipulation, simulator
\end{IEEEkeywords}

\section{Introduction}
Rock skipping has been one of the most famous pastimes for thousands of years with many people today having fond childhood memories of throwing stones across a lake. Even for kids, rock skipping is a task that can easily be accomplished. However, the act of rock skipping itself is a highly dynamic and relatively complex task. After taking Robotic Manipulation, we were motivated to see if we can try to replicate this task in a robotic setting. Specifically, we wanted to see if we could design and implement a system in simulation that utilizes an IIWA arm to complete the task of rock skipping. This task can be broken down to several components: picking up the rock, throwing it, and then simulating the water skipping dynamics. Ultimately, by creating a system in simulation, we hoped to vary parameters such as release height and velocity in order to gain insight into what factors were most important for maximizing the total number of skips. In addition to gaining insight into rock skipping, we were motivated to learn how to implement a dynamic system in simulation. 

The overall literature on rock skipping is pretty minimal, especially any literature in a robotic setting. As a result, the related works will be discussed in the introduction. One of the most influential papers we used for our project is a study from Truscott Et al. \cite{Truscott_Belden_Hurd_2014} which analyzes the physics behind the phenomenon of skipping. The paper not only provides the force relations when the rock is in contact with the water, but also provides great intuition for how differing setups affect performance. This was extremely useful for our project as the paper gave us an intuitive understanding of the potential effect of the parameters. The second related work is from Mark Rober's Youtube video on his own personal rock skipping robot \cite{youtube_rock_skipping}. Taking a very experimental approach, Mark concluded that rock angle relative to water, spin of the rock, and rock shape were the most important factors. However, this experimental setup was limited in the sense that the robotic arm used was simply a renovated clay pigeon thrower. This means the video lacked precision in the measurements taken due to the real-world nature of the experiment, any form of trajectory optimization, and automation such as picking up the rock. We hope our project will be able to fill in some of the gaps with a more rigorous simulated approach.  

\section{Methods}
\subsection{Simulator Set-Up}
\begin{figure}[htbp]
\centerline{\includegraphics[width=5cm]{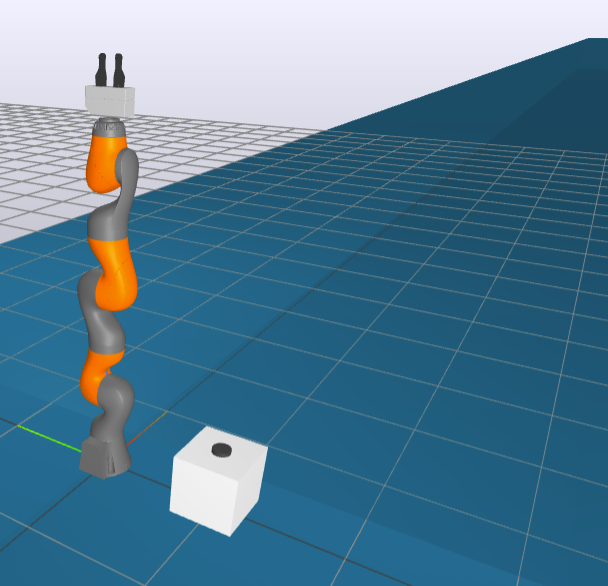}}
\caption{Image of our simulation setup with the IIWA arm, gripper, table, water, and rock.}
\label{fig:sim_setup}
\end{figure}

The first step to our project was setting up our simulation. We decided to use PyDrake \cite{drake} as the foundation for our project since we have been using it all year in class. Within PyDrake, we utilized the manipulation station which consists of a Kuka IIWA LWR for our robotic arm and a simple two-finger Schunk WSG 50 for our gripper. We figured the IIWA arm and Schunk gripper provided enough flexibility for our rock skipping task. In addition to choosing the manipulation station, we created model instances using primitive geometries in SDFormat for the water, table, and rock objects. These primitive geometries are parsed by drake to create the model in simulation. The entire setup is displayed below in \Cref{fig:sim_setup}. It is important to note that we assume an abundant number of camera perspectives for this project and known location of objects. 

We had to make design choices when creating the models for the water, table, and rock. Both the water and table were relatively easy choices since any reasonable choice would have no significant impact on the rock skipping task. We simply placed them in locations advantageous for our manipulation task. The water is a long rectangular model in the positive x direction with a height at 0 m so its ground level. The table is a box shape with its most important parameter being its height. On the other hand, the design of the rock has a significant impact on the performance of our system. This conclusion was reached in both of the previous works \cite{youtube_rock_skipping, Truscott_Belden_Hurd_2014} as both works found that a flat rock geometry resulted in more skips. This will be explained more in the context of the skipping dynamics in the next section. As a result, we choose to make our rock resemble a flat hockey puck shape. This shape matches the flat rock geometry requirement and also provides a large top and bottom surface area for the gripper. The most important parameters and its values for the system's models are shown in \Cref{table:system_parameters}.  

\begin{table}[htbp]
\caption{System Parameters}
\begin{center}
\begin{tabular}{|P{2.2cm}|P{2.2cm}|P{2.2cm}|}
\hline
Rock radius & 0.035 m \\ \hline
Rock length & 0.015 m \\ \hline
Rock mass &  0.25 kg \\ \hline
Table height & 0.25 m \\ \hline
Water height & 0 m \\ \hline
\end{tabular}
\label{table:system_parameters}
\end{center}
\end{table}

Lastly, our simulation setup includes a force system controller to apply the spatial forces from the water onto the rock. Collision geometry of the water is ignored and all impact forces are computed according to a later described model. The force system takes in the calculated water force and converts it to an ExternallyAppliedSpatialForce type in PyDrake. This type enables us to indicate what body in the simulation the force is applied to. In order to utilize this output from the force system, we had to expose the internal spatial force port within the station. This is accomplished during the creation of the station by exporting the internal spatial force port as an input port.

\subsection{Skipping Dynamics}

\begin{figure}[htbp]
\centerline{\includegraphics[width=6.5cm]{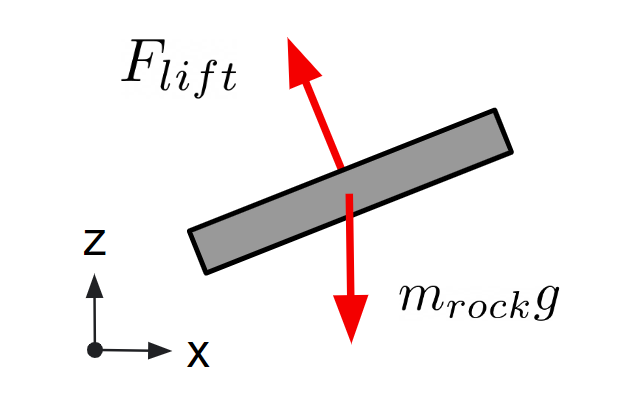}}
\caption{Free body diagram of a rock colliding with water surface. Upward force from water surface defined as $F_{lift}$.}
\label{fig:fbd}
\end{figure}

Rock skipping occurs when an upward force on the surface of water counteracts the force of gravity. We define this force as $F_{lift}$. If executed properly, this creates an upward acceleration to bounce the rock off of the water and back into the air. The free-body diagram of a flat rock colliding with the water displaying $F_{lift}$ is shown in \Cref{fig:fbd}.

Upward force $F_{lift}$ can be modeled as a drag force. Previous papers \cite{Truscott_Belden_Hurd_2014} have employed this technique and used \Cref{eqn:lift}.

\begin{equation}\label{eqn:lift}
F_{lift} = \rho_{water}C_DS_{wet}||\bar{V}||^2sin(\beta + \alpha)\hat{n}
\end{equation}

The drag force acts in the upward normal direction $\hat{n}$. It is directly proportional to a coefficient of drag $C_D$. For our use case, we will take this coefficient to be equal to 0.5. The force is also proportional to the density of the liquid it is skipping over $\rho_{water}$. This makes sense because if the liquid is more dense, the rock is more likely to bounce off of it. 

\begin{figure}[htbp]
\centerline{\includegraphics[width=6.5cm]{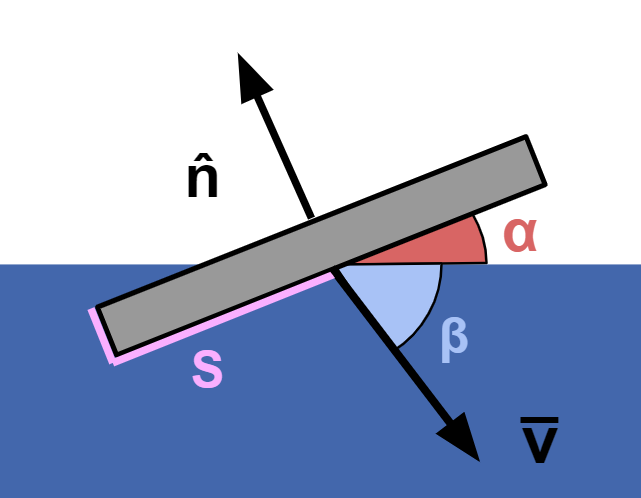}}
\caption{Visually defined parameters of rock skipping. $\hat{n}$ is the upward vector normal to the rock's surface. $S$ is the wetted area of the rock. $\alpha$ is the angle between the rock and water surfaces. $\bar{V}$ is the planar velocity of the rock in three dimensions. $\beta$ is the angle between $\bar{V}$ and water surface.}
\label{fig:forceparam}
\end{figure}

The force is proportional to the wetted area of the rock $S_{wet}$. As you increase the contact area between the rock and water, you get a higher upward force. This is intuitive. It also implies that a very flat rock geometry will give us more skips because it can achieve a larger wetted area for a given mass. Parameters of $F_{lift}$ are displayed in \Cref{fig:forceparam}.

The angle $\alpha$ is defined as the angle between the surface of the rock and the surface of the water. This measures the flatness of the rock on impact. It can be calculated according to \Cref{eqn:alpha}.

\begin{equation}\label{eqn:alpha}
\alpha = acos(\hat{n}\cdot \hat{x}) - \pi/2
\end{equation}

The angle $\beta$ is defined as the angle between the rock's planar velocity $\bar{V}$ and the surface of the water. This measures the flatness of the rock's velocity on impact. It can be calculated according to \Cref{eqn:beta}.

\begin{equation}\label{eqn:beta}
\beta = \pi/2 - acos(\frac{-V_z}{||\bar{V}||})
\end{equation}

The upward skipping force is proportional to the vector norm of the rock's planar velocity $\bar{V}$. This tells us that the faster you throw a rock, the more likely it is to skip. However, in accordance with angle $\beta$, this velocity should be as flat as possible with minimum element in the $\hat{z}$ direction.

We can take advantage of these force model parameters in planning to create an advantageous system for rock skipping.

Outside of the lift force $F_{lift}$, we also supply a damping force on impact to account for frictional losses. This force is defined in \Cref{eqn:damping} and is intended to slow the rock down as it skips. This should increase the accuracy of our simulation. Damping coefficient $D$ is chosen based on prior experimental studies\cite{EscalanteMartnez2016}.

\begin{equation}\label{eqn:damping}
F_{damping} = -D\bar{V}
\end{equation}

With $F_{damping}$ defined, we simply add this to $F_{lift}$ to get the total force $F_{rock}$ that should be supplied to the rock on surface impact.

\begin{equation}\label{eqn:F}
F_{rock} = F_{lift} + F_{damping}
\end{equation}

\subsection{Planning}

\begin{figure}[ht]
\centering
\begin{subfigure}{\linewidth}
    \centering
    \includegraphics[width=.5\linewidth]{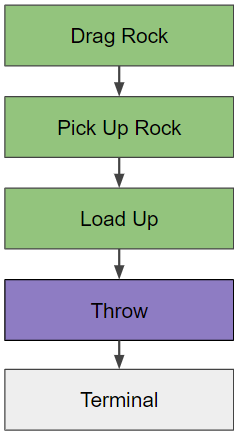}
\end{subfigure}
\begin{subfigure}{1.35\linewidth}
    \vspace{0.25cm}
    \hspace{0.105\linewidth}
    \includegraphics[width=.5\linewidth]{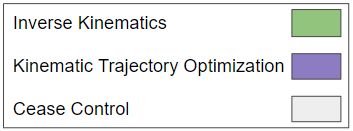}
\end{subfigure}
\caption{The planning state diagram shows the 5 different states for our robotic arm. The color of each state reveals the control method used as displayed by the bottom legend.}
\label{fig:Planner_State_Diagram}
\end{figure}

The task of rock skipping can be broken down into 4 main stages: bringing the rock to an ideal pickup position, picking it up, loading up the rock before throwing, and actually throwing the rock. Since our robotic arm is required to go through these different stages, we implemented a state machine to handle the planning of the arm. By utilizing a state machine, we can easily switch the way we command and control the arm. These different stages and the underlying control methods used to command the arm are shown in \Cref{fig:Planner_State_Diagram}. It's important to note that we have a terminal state where the arm is no longer commanded by anything. We enter this state while we are simulating the rock skipping after it has been thrown.  

\begin{figure}[ht]
\centerline{\includegraphics[width=7cm]{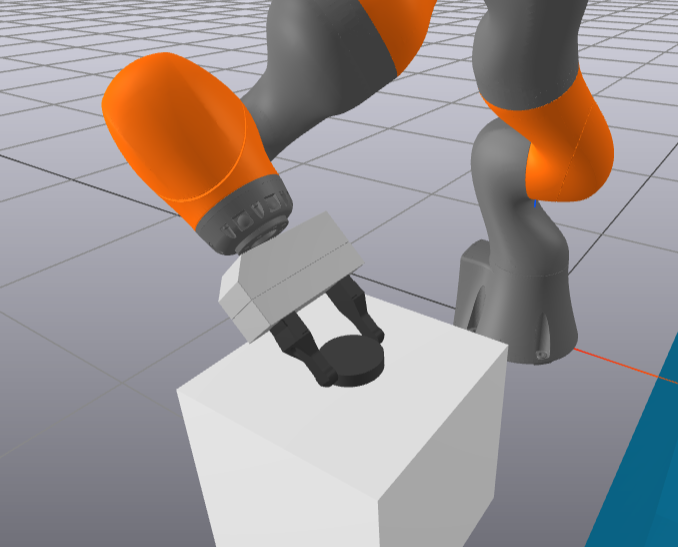}}
\caption{Manipulation of the rock to reposition it from the center of the table to the edge. By dragging the rock to the edge of the table, we can utilize a more ideal grasp position on the rock.}
\label{fig:prepick_rock}
\end{figure}

Since the rock starts at the center of the table, the first state is designed to get the rock to a better position to pick it up. It is important to restate that we assume perfect perception with an abundant number of camera perspectives and a known rock location for this project. We manipulate the rock by dragging it to the edge of the table as seen in \Cref{fig:prepick_rock}. In order to drag the rock, we used simple kinematic planning and differential inverse kinematics as the underlying control method for the arm. Specifically, we created a sequence of desired end-effector poses and specified the difference of time $\Delta t$ between each pose. Differential inverse kinematics is then used to determine the necessary joint commands to achieve those poses. We drag the rock to the edge of the table in order to get a strong antipodal grasp with the gripper grasping the two large flat surfaces of the rock (top and bottom). 

\begin{figure}[ht]
\centerline{\includegraphics[width=7cm]{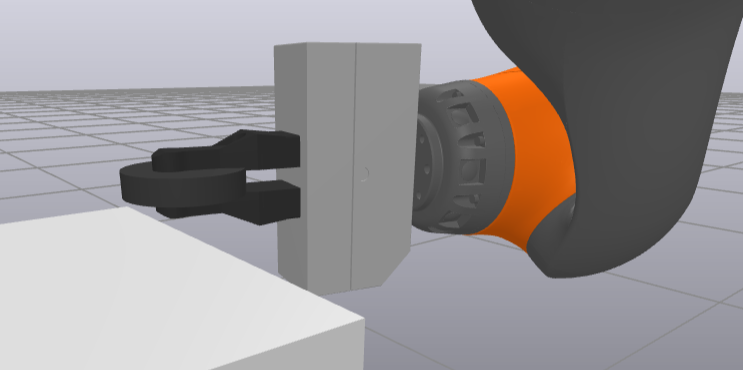}}
\caption{The gripper clamps the rock in a similar manner as a human would hold a frisbee. This antipodal grasp maximizes the contact area between the rock and the fingers of the gripper which allows for a more secure hold.}
\label{fig:pick_up_state}
\end{figure}

After dragging the rock, we enter the second state which is picking up the rock. The antipodal grasp mentioned before is shown in \Cref{fig:pick_up_state}. Similarly to the dragging state, we continue to use simple kinematic planning and differential inverse kinematics during this state to command the arm. The third state is loading up the rock to throw. In order to gain the most speed during the throw, we position the arm the farthest away from the water we can. The motion planning and control method is the same as the two other states. 

The final state that requires control of the arm is the throwing state. The exact optimization is mentioned explicitly below in the Throwing Optimization section. Ultimately, we utilize Kinematic Trajectory Optimization as the underlying control method within this state. This control method is required to properly consider velocity constraints that could not be accomplished with a simple inverse kinematics approach.

\subsection{Throwing Optimization}

Once the rock has been gripped, we must throw it at a proper speed to cause skipping. During pick-up, we grab the rock in such a way to take advantage of the rock's cylinder geometry. By clamping down on the rock's flat sides, we have an antipodal grasp that can effectively hold the rock with high normal force throughout the duration of the throw until release.

The objective of rock skipping is to throw the rock at a desired velocity from a desired release point. Originally, we attempted to throw the rock using a simple differential inverse kinematics approach. This is the same approach we used to plan the pickup of the rock. We commanded a set of desired end-effector poses wit varying $\Delta t$ between them that speed up to a desired velocity until the rock's release. However, this approach did not work. Differential inverse kinematics does not fully consider the set of desired poses, it only maps a path between sequential desired poses. Therefore, this approach has a difficulty achieving the high velocities required to skip a rock. It may successfully follow the first few desired poses, but eventually it's joints may reach an unadvantageous position that restricts the solver from planning paths to future poses. This is compounded by the fact that later poses are required to reach a higher speeds across the throw. After this approach failed, we had to find a more comprehensive approach that planned across the whole path and had a better intuition of desired velocity.

\begin{figure}[htbp]
\centerline{\includegraphics[width=7cm]{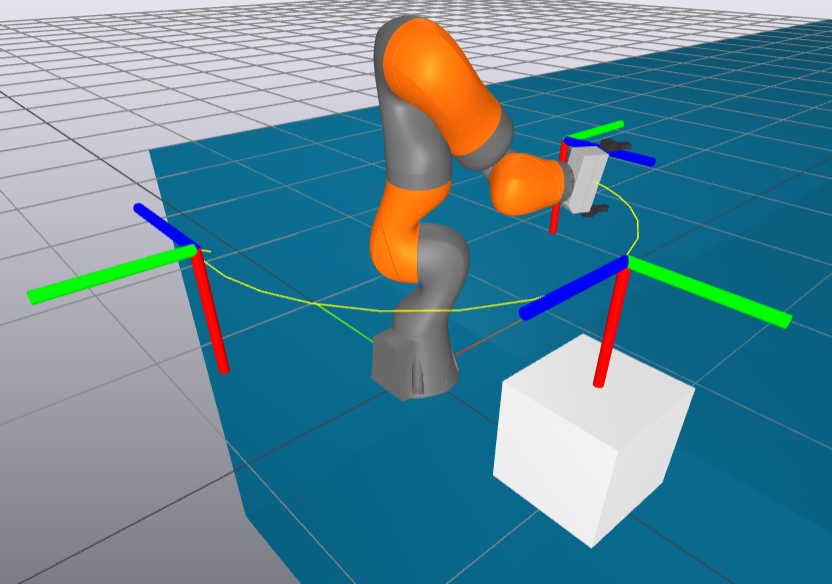}}
\caption{Trace of trajectory path with desired release velocity 25 m/s and desired release height of 0.5m. Initial $p_0$, release $p_r$, and final $p_f$ constraint poses are shown with respective coordinate frames.}
\label{fig:throw_traj_visual}
\end{figure}

Instead, we used kinematic trajectory optimization to find the optimal path for throwing at a desired velocity from a desired release point. This approach considers all constraints when finding the optimal, comprehensive throwing trajectory. From a high-level view, we specify an initial load-up position $p_0$, a release position $p_r$, and an end position of the end-effector to follow through to after release $p_f$. These positions follow an approximately radial path about the origin of the robot frame. The trajectory path and constraint poses can be seen in \Cref{fig:throw_traj_visual}.

The trajectory optimization formulation is outlined in \Cref{eqn:trajopt}. We impose loose position constraints of the end-effector around the three specified poses with error bound $\epsilon_p$. These create a box around each that the gripper must pass through. We impose more strict constraints around the $\hat{z}$ coordinate of the release position. This is because the release height has a great effect on the resulting trajectory of the rock. We do not care as much about the rock's release point along the $\hat{x}$ and $\hat{y}$ axes.

Similarly, we impose an orientation constraint with very small angular bound $\epsilon_\theta$ at the release point to make sure that the rock is released as flat as possible. More rigorously, the normal force of the rock should be parallel to the $\hat{z}$ axis.

Most importantly, we impose a constraint on spatial velocity with small error bound $\epsilon_v$ at the release point. This requires that our end-effector is moving at an input desired $\hat{x}$ velocity. Once again, this is highly important throwing because we want to make sure the rock is thrown at a velocity that is advantageous to allow for skipping.

\begin{equation}\label{eqn:trajopt}
\begin{aligned}
\min_{q} \quad & T_{traj} + L_{traj} \\
\textrm{s.t.} \quad & \text{ForwardKinematics}(q_i) = p_i\\
  &p_{0,des} - \epsilon_{p} \leq p_{0} \leq p_{0,des} + \epsilon_{p}\\
  &p_{f,des} - \epsilon_{p} \leq p_{f} \leq p_{f,des} + \epsilon_{p}\\
  &p_{r,des} - \frac{\epsilon_{p}}{4} \leq p_{r} \leq p_{r,des} + \frac{\epsilon_{p}}{4}    \\
  &R_{r,des} - \epsilon_{\theta} \leq R_{r} \leq R_{r,des} + \epsilon_{\theta} \\
  &\dot{p}_{r,des} - \epsilon_{v} \leq \dot{p}_{r} \leq \dot{p}_{r,des} + \epsilon_{v}    \\
\end{aligned}
\end{equation}

We minimize the length and duration of the throwing trajectory to get a smooth and fast resulting trajectory. The optimization problem is solved using the SNOPT solver \cite{doi:10.1137/S0036144504446096} implemented in PyDrake.

\section{Results \& Discussion}

\subsection{Release Velocity}

The dynamics model we employed to model skipping implies that skipping lift force is highly dependent on velocity. We ran our planning structure on a set of desired velocities to measure the effect that velocity has on skipping. The robotic system did not perfectly achieve the desired release velocities. This is likely due to imperfections on throw release. Our two-finger gripper is sub-optimal for throwing and cannot release the rock as directly as hoped. The rock may collide with the gripper on release and decelerate. Discrepancies between desired velocities and actual release velocities are tabulated in Table II.

\begin{figure}[htbp]
\centerline{\includegraphics[width=8cm]{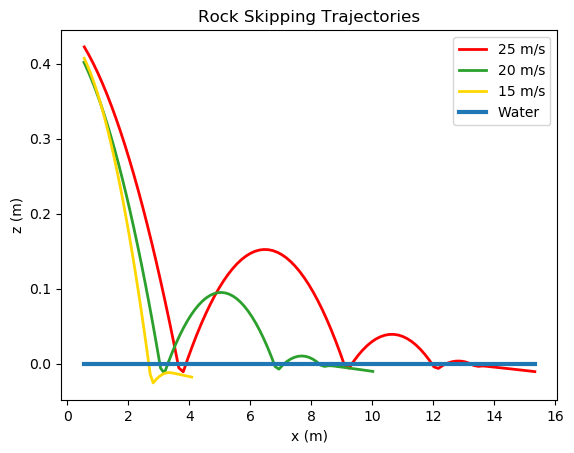}}
\caption{Resulted skipping trajectories for a set of commanded desired output velocities $\hat{x}_{des}$. Surface of water shown in blue.}
\label{fig:skipping_traj}
\end{figure}

Despite this fact, rock velocities were still quite high and able to skip across the water in simulation. The resulting trajectories for respective desired throwing velocities are graphed in \Cref{fig:skipping_traj}.

\begin{table}[htbp]
\caption{Output Throwing Velocities}
\begin{center}
\begin{tabular}{|P{1.8cm}|P{1.8cm}|P{1.8cm}|}
\hline
Desired throw velocity ($\dot{x}$) & Output rock velocity ($\dot{x}$) & Number of Skips \\

\hline
25 m/s & 14.4 m/s & 4 \\ 
\hline
20 m/s & 12.6 m/s & 3 \\ 
\hline
15 m/s & 9.9 m/s & 0\\ 
\hline
\end{tabular}
\label{tab1}
\end{center}
\end{table}

We can conclude a few things from these resulting trajectories. Firstly, it is clear that a higher velocity $\dot{x}$ leads to a longer trajectory length and greater number of skips. We can clearly see that the rock thrown with highest desired velocity achieves the most skips. The rock thrown at a lower velocity still skips, but does not skip as many times or go as far.

Furthermore, we can conclude that a rock must be thrown above a certain a velocity to actually skip. All other conditions withstanding, there is effectively a cut-off velocity which must be superseded in order to counteract the force of gravity. We see with the rock thrown at desired velocity of 15 m/s, it does not skip. This rock is thrown at an actual velocity of 9.9 m/s. So, we expect that the rock in our system must be thrown at at least 10 m/s to allow for skipping.

Many limitations were faced in attempt to get the IIWA arm to throw the rock at these high velocities. We only ran test trials below 25 m/s because any desired velocity over this led to slipping. When slipping occurred, the rock would preemptively fly out of the gripper before reaching the prescribed release point. We grasped the rock using the best possible anti-podal grasp on the flat sides of the cylinder, but this was not enough. We attempted a handful of preventative measures to solve this problem. First, we maximized the normal force of the grasp. Slipping still occurred. Second, we tried to increase the friction of the rock-gripper interaction. When that didn't work, we tried to decrease the $\Delta t$ of our simulator. None of these methods allowed us to properly throw above 25 m/s. However, they did help us reach the 25 m/s maximum desired velocity that we observed. It is a lot to ask the simulator to work at velocities above this threshold. In the future, we would hope to further fine-tune the above mentioned parameters and likely design a new gripper. A customized multi-finger gripper that is more optimized for rock skipping would greatly increase our chances of throwing above 25 m/s and avoid premature slipping.

\subsection{Release Height}

Our dynamics model tells us that a flatter velocity with minimum $v_z$ element is best for skipping. We hoped to test this hypothesis by running our implementation on a set of varying desired release heights. Our trajectory optimizer constrains that the rock must be released at nearly the exact release height. However, the solver had difficulty finding solutions for many different release heights. This meant we could not run as many trials with varying release heights as we had initially hoped to. We suspect that the reason the trajectory optimization did not work at different release points is because we had fine-tuned our setup to operate at a specific release height of 0.5m.

\begin{figure}[htbp]
\centerline{\includegraphics[width=8cm]{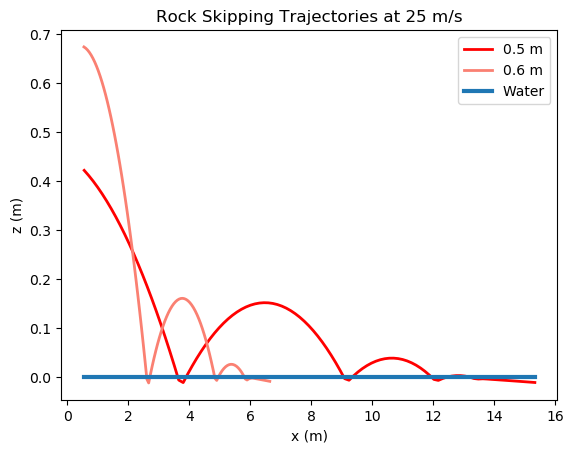}}
\caption{Resulted skipping trajectories for a set of commanded desired output velocities $\hat{x}_{des}$. Surface of water shown in blue.}
\label{fig:release_height}
\end{figure}

Despite difficulty, we still did get some successfully solved trajectories at a different release height. Results comparing these different release heights at a constant throwing velocity are plotted in \Cref{fig:release_height}. It is clear that the higher release height only creates 3 skips. This is less than the lower release heights, which skips 4 times. This matches our assumption from the dynamic model of skipping we use.

Due to problems with solving for trajectories, we cannot fully prove that a lower release height is best for rock skipping. However, the two trajectories we have generated suggest that a lower release would likely lead to less skips. In the future, we would need to further tune our planning system in order to get more reliable data.

\section{Conclusion}

Overall, our system was able to demonstrate skipping and generate some preliminary results. However, it has a fair number of limitations. We showed that a higher throw velocity leads to more rock skips. But, we could not show this over a wide range of velocities due to gripping inefficiencies. We also suspect that a lower release height gives more skips based on preliminary results. This cannot be rigorously proven due to issues with trajectory optimization.

In the future, we would like to improve the capabilities of our system. Most importantly, we would like to create a customized gripper with geometry that prevents the rock from prematurely slipping even at high velocity throws. This way we can test over a wider range of velocities and hopefully get a more impressive number of rock skips. We would also like to generalize our approach to trajectory optimize to test over a more diverse set of release points. These improvements would enable us to have a more holistic investigation of the effect that throwing has on rock skipping.

Our system assumed a constant rock geometry at a known location. Rock geometry certainly has an effect on skipping. We chose our rock shape to be flat and cylindrical because this is what is typically sought after by real-world rock skippers. A more generalized planning system would allow us to test a wide variety of rock geometry and outline a clearer picture of which rock shapes are best. This type of study would allow us to develop a perception system which can choose the best rock to skip over a set of choices. Implementing this into our system would make the system capable of skipping a rock autonomously in a real-world like setting.

\section{Contributions}

In terms of contributions, our work overlapped significantly as we both worked on tasks such as simulation setup, overall implementation, and  debugging together. As a result, the division of work isn't too distinctly separable. However, there were two sections of our project that were implemented entirely by a single team member. Michael designed and implemented the skipping dynamics, and Nicholas created the planner state machine.

\printbibliography

\end{document}